\documentclass[10pt,twocolumn,letterpaper]{article}

\usepackage{cvpr}              

\usepackage{graphicx}
\usepackage{amsmath}
\usepackage{amssymb}
\usepackage{booktabs}

\usepackage[pagebackref,breaklinks,colorlinks]{hyperref}

\usepackage[capitalize]{cleveref}
\crefname{section}{Sec.}{Secs.}
\Crefname{section}{Section}{Sections}
\Crefname{table}{Table}{Tables}
\crefname{table}{Tab.}{Tabs.}

\def\cvprPaperID{*****} 
\def\confName{CVPR}
\def\confYear{2023}

\begin{document}

\title{SRCN3D: Sparse R-CNN 3D for Compact Convolutional Multi-View 3D Object Detection and Tracking}

\author{
  Yining Shi$^{1}$, Jingyan Shen$^{1}$,  Yifan Sun$^{1}$, Yunlong Wang$^{1}$, \\ Jiaxin Li$^{1}$, Shiqi Sun$^{1}$, Kun Jiang$^{1}$\footnotemark[2], Diange Yang$^{1}$\footnotemark[2] \\
  $^{1}$ Tsinghua University
}

\maketitle
\renewcommand{\thefootnote}{\fnsymbol{footnote}}
\footnotetext[2]{Corresponding author: Kun Jiang, Diange Yang}

\begin{abstract}
Detection and tracking of moving objects is an essential component in environmental perception for autonomous driving. In the flourishing field of multi-view 3D camera-based detectors, different transformer-based pipelines are designed to learn queries in 3D space from 2D feature maps of perspective views, but the dominant dense BEV query mechanism is computationally inefficient. This paper proposes Sparse R-CNN 3D (SRCN3D), a novel two-stage fully-sparse detector that incorporates sparse queries, sparse attention with box-wise sampling, and sparse prediction. SRCN3D adopts a cascade structure with the twin-track update of both a fixed number of query boxes and latent query features. Our novel sparse feature sampling module only utilizes local 2D region of interest (RoI) features calculated by the projection of 3D query boxes for further box refinement, leading to a fully-convolutional and deployment-friendly pipeline. For multi-object tracking, motion features, query features and RoI features are comprehensively utilized in multi-hypotheses data association. Extensive experiments on nuScenes dataset demonstrate that SRCN3D achieves competitive performance in both 3D object detection and multi-object tracking tasks, while also exhibiting superior efficiency compared to transformer-based methods. Code and models are available at \url{https://github.com/synsin0/SRCN3D}.
\end{abstract}

\section{Introduction}
\begin{figure*}[htbp]
    \centering
    \includegraphics[width=1\textwidth]{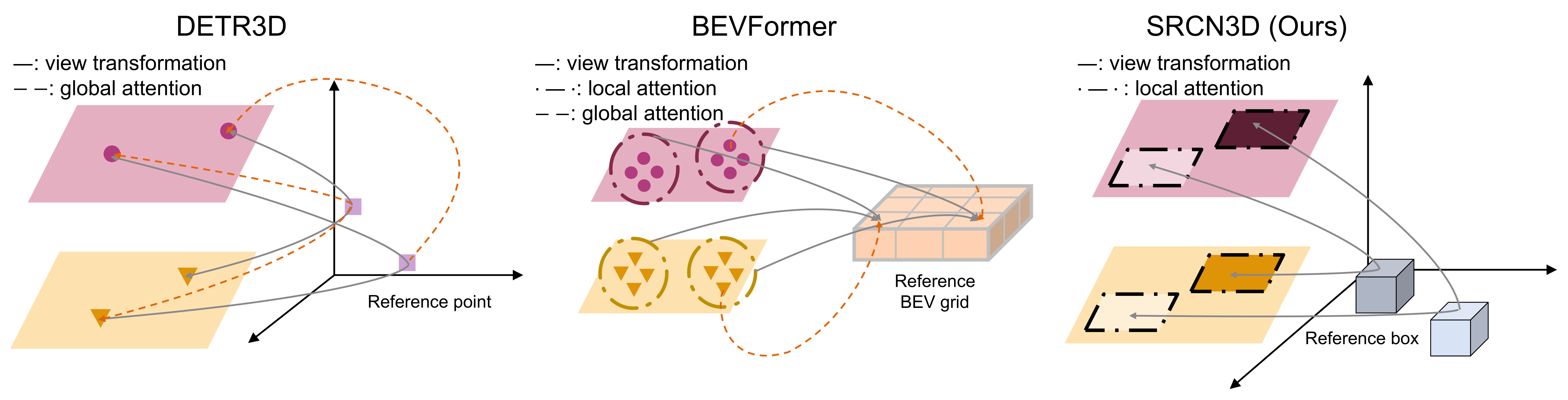}
    \caption{Comparison between typical query-based multi-view 3D (MV3D) detectors. DETR3D \protect\cite{DETR3D} applies sparse queries and dense global attention. BEVFormer \protect\cite{BEVFormer} applies dense queries and dense deformable attention. SRCN3D steps further to sparser queries and totally sparse local RoI attention, which pioneers a fully-sparse box-wise sampling paradigm.}
    \label{comparison}
\end{figure*}

\noindent Environmental perception is an essential task in the field of autonomous driving. 3D object detection and tracking are responsible for identifying and localizing objects of interest, as well as recording their unique labels and past trajectories. While LiDAR is commonly used for this purpose, on-board cameras offer certain advantages such as lower cost, wider detection range, higher angular resolution, and richer semantic cues. However, vision-centric detectors face two long-standing challenges. Firstly, since cameras lack geometric or depth cues, 3D reconstruction can be an ill-conditioned problem. Secondly, there is the issue of cross-view fusion, which refers to the challenge of detecting an object as a whole when two adjacent cameras only capture parts of it.   

Vision-centric detectors rise rapidly in accuracy thanks to latest innovation of data-driven view transformation from perspective view to 3D world space. LSS \cite{LSS} proposes a geometry-based explicit pipeline that includes depth estimation, point-cloud lifting, voxelization, and splatting, while DETR3D\cite{DETR3D} introduces a network-based implicit pipeline learning queries from projected reference points to obtain values without explicit depth estimation and post-processing. As illustrated in Fig. \ref{comparison}, previous arts require either dense queries or dense interaction between queries and values. Our proposed method enjoys the merit of implicit pipeline and avoids dense feature sampling at the same time.

Our proposed method, SRCN3D provides a simple and elegant cascade pipeline and set prediction approach highly inspired by Sparse R-CNN \cite{SRCN}. We adopt commonly used backbones and Feature Pyramid Network (FPN) \cite{FPN} neck, but a novel SRCN3D head that iteratively updates both 3D query boxes and query features at the same time. Specifically, 3D query boxes are projected to six views to aggregate local RoI features. The resulting features are used to refine query features via a sparse interaction head, which produces classification and the offsets that are applied to the original bounding boxes. After three to six stages, the final refined query boxes serve as the direct detection outputs, eliminating the need for complicated post-processing steps or regression processes from latent features. Our framework does not rely on transformer-style operations such as masking operations or positional embeddings.

Besides, in a mainstream tracking-by-detection (TBD) pipeline, multi-object tracking is challenging in the data association process given possible missed or false detection. This problem is more common for camera-based detectors than LiDAR-based detectors. Therefore, conventional deterministic data association suffers from poor track continuity. Recent trackers, such as QD-3DT \cite{QD-3DT}, incorporate dense image features, while re-identification (Re-ID) from dense region queries is inefficient and inaccurate. In light of the instability of detection results, we find out that a simplified multi-hypotheses Random Finite Set (RFS) approach for probabilistic matching reduces failed tracking. Our proposed tracker is also the first approach to incorporate RoI features and query features in the RFS framework.     

In summary, this paper makes the following contributions:

\begin{itemize}
\item To the best of our knowledge, \textbf{the first transformer-less two-stage MV3D approach with box-wise sampling.} The pipeline is straightforward, lightweight, and faster than other transformer-based detectors.  

\item \textbf{A novel sparse cross-attention module to refine 3D queries from 2D feature maps}, which replaces dense attention with a local sparse interaction module. Consequently, a lower computation cost is achieved.

\item  Extensive experiments on nuScenes dataset demonstrate the effectiveness of SRCN3D for 3D object detection and tracking.
\end{itemize}

\section{Related Work}

\begin{figure*}[htbp]
\centering
\includegraphics[width=1\textwidth]{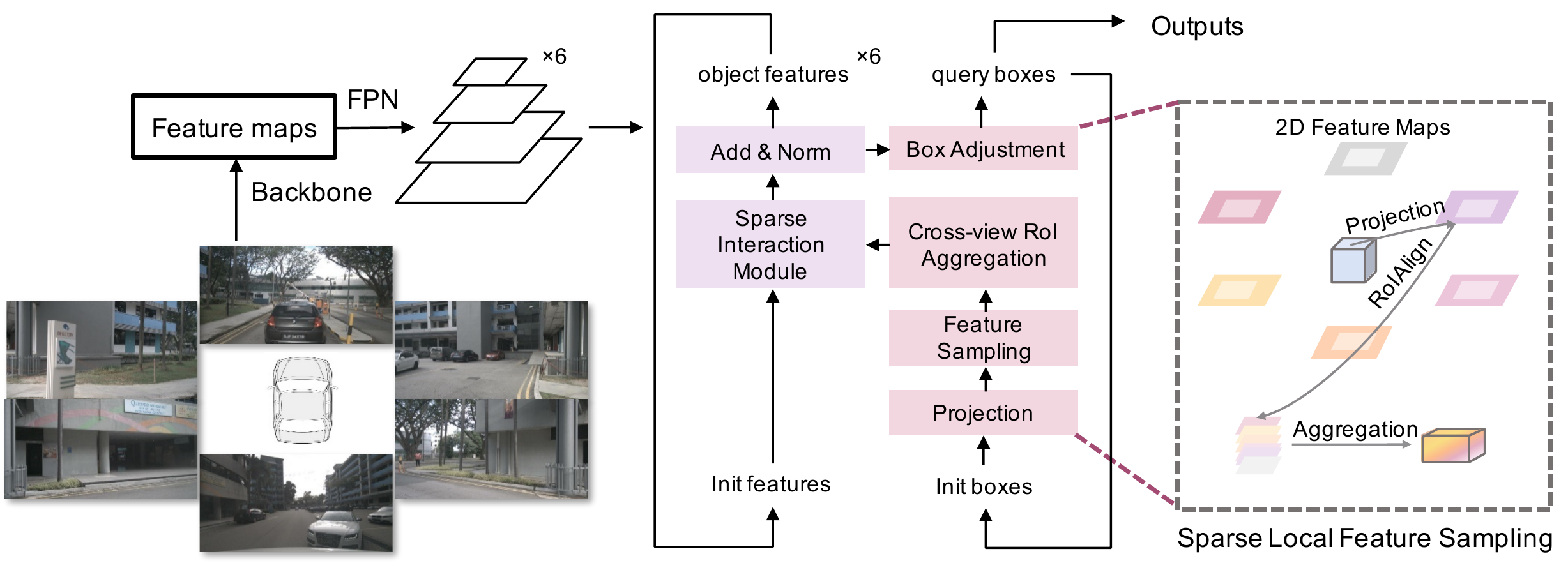}

\caption{The framework of SRCN3D. Taken camera images as inputs, SRCN3D contains a backbone network with FPN to extract 2D feature maps and a twin-track detection module. A sparse feature sampling module is designed to extract local RoI features and refine the query boxes.}
\label{framework}
\end{figure*}

\subsection{Multi-camera 3D Object Detection} The development of MV3D has progressed rapidly due to the use of data-driven view transformation. Currently, there are three primary paradigms for state-of-the-art multi-view camera 3D object detection: perspective-view-based (PV-based), geometry-based, and network-based view transformation. PV-based view transformation involves predicting 3D boxes from 2D images in perspective view and performing BEV aggregation \cite{FCOS3D,PGD,MonoDETR,DD3D}, which presents challenges in cross-view fusion. Geometry-based view transformation involves depth estimation, followed by transformation to world coordinates and aggregation of point-clouds in a bottom-up manner. However, this approach is sensitive to depth errors. Network-based view transformation, on the other hand, involves randomly proposing 3D queries and refining them in a cascade \cite{DETR3D,PETR,BEVFormer}, without depth estimation. Our proposed method adopts the network-based paradigm and introduces a novel two-stage pipeline that differs from previous single-stage approaches.

\noindent 

\subsection{Transformer-based Object Detection} 
Vision transformers have demonstrated excellent performance in object detection. In 2D domain, DETR \cite{DETR} presents a set-prediction paradigm with regard to a fixed set of objects. Deformable DETR \cite{Deformable-DETR} invents deformable attention for faster convergence. Furthermore, Sparse R-CNN \cite{SRCN} puts forward a completely sparse schema, where each query box interacts only with its specific query feature. In 3D domain, transformers serve as cross-attention between 3D queries and 2D feature maps\cite{DETR3D,BEVFormer,PETR,PETRv2}. DETR3D \cite{DETR3D} is the first to apply a top-down framework, which projects reference points on feature maps and performs cross-attention to refine query features. BEVFormer \cite{BEVFormer} leverages cross-attention on bird's eye view (BEV) grid features, and employs spatial deformable attention for BEV grids and temporal alignment of BEV features. PETR \cite{PETR} and PETRv2 \cite{PETRv2} adopt ideas from implicit neural representation and project 2D features maps to 3D space so as to interact with 3D queries. Compared to dense attention arts, our method explores a sparse feature sampling module without global attention and makes itself a purely sparse approach. 

\noindent \subsection{3D Multi-Object Tracking} 3D Multi-Object Tracking (3D MOT) is another challenging task right after object detection, aiming to temporally associate trajectories of each same object and record its unique label. Data association is the core issue of MOT, where the dominant matching approach is Global Nearest Neighbor (GNN), carried out in the form of Hungarian or Greedy algorithm used in AB3DMOT \cite{AB3DMOT}. Another approach is RFS \cite{PMBM_deri}, an online multi-hypotheses paradigm circumventing deterministic one-to-one pair matching. Current MOT researches, e.g. CenterTrack \cite{CenterTrack}, QD-3DT \cite{QD-3DT} and MUTR3D \cite{MUTR3D}, focus on utilizing implicit features to express matching similarity in the embedded space, empowering the network to identify the same object based on appearance features through Re-ID process. Our tracker explores a novel multi-hypotheses probabilistic mode of data associated with hybrid feature embedding.

\section{Methodology}


\subsection{Overview}

The overall framework of SRCN3D is illustrated in Fig. \ref{framework}. SRCN3D has a novel two-stage fully-convolutional 3D object detection architecture. We build our architecture based on the following consideration:
\begin{itemize}
    \item We estimate 3D bounding boxes directly in 3D world space without depth supervision and post-processing like non-maximum suppression (NMS), while 2D images serve as implicit cues in our network. 
    
    \item Following a sparse paradigm, each query box serves as a filter to focus on a sparse local region of 2D feature maps.
    
    \item We design a fully-convolutional pipeline without mask operations, positional embeddings and attention weights in typical vision transformers.
    
\end{itemize}

The architecture of SRCN3D consists of a common backbone with FPN and a novel SRCN3D head. First of all, we feed the RGB images into a backbone network (e.g. ResNet-101 \cite{Resnet}) with FPN \cite{FPN} to generate multi-level multi-camera feature maps $\{F_{1}^{i},F_{2}^{i},F_{3}^{i},F_{4}^{i}\}_{i=1}^{N_{cam}}$ for each view. SRCN3D heads start from a fixed set of 3D learnable query boxes and query features, and the initial parameters for boxes and features are randomly initialized and learnable during training process. In a cascade structure, the query boxes and query features compose strict pairs while non-pairs do not interact. Boxes and features are updated in a twin-track approach. Query boxes are updated in a box adjustment step and query features are updated with a sparse feature sampling module. The last stage of query boxes and class regression of query features without post-processing techniques comprise the final results.

\subsection{SRCN3D Head}

As shown in Fig. \ref{sparse_feature_sampling}, SRCN3D head composes of three key modules, that is, a sparse feature sampling module for feature extraction, a sparse interaction module for feature refinement and a box adjustment module for box refinement. For each iteration round, there are two inputs: a fixed number of learnable query boxes and query features. The forward process for each round includes the following steps:
\begin{itemize}
    \item Restore query boxes to world scale, retrieve eight corner points and project corner points to images using corresponding intrinsic and extrinsic camera parameters.
    \item Collect four borderline corner points from each image to form a RoI candidate, sample RoI features using RoIAlign and aggregate cross-view RoI features.
    \item Feed cross-view RoI features into sparse interaction head to refine query features and output cues (fine-tuning values) for box adjustment.
    \item Refine location, dimension, rotation and velocity with box adjustment module, normalize the 3D positions of the boxes and obtain the input for the next round.
\end{itemize}

\textbf{3D query boxes} are defined as a fixed number of boxes parameterized to the same dimension as 3D bounding box (e.g. $\{B_i\}_{i=1}^{N} \subset R^{10}, N = 300$). The 10 dimensions are defined as $[c_x,c_y,h,w,c_z,l,cos\theta,sin\theta, v_x,v_y]$, where $c_x,c_y,c_z$ are center coordinates of the box, $h,w,l$ are height, weight and length, $\theta$ is the yaw angle and $v_x,v_y$ are velocities.

\textbf{Query features} are represented by sets of high-dimensional latent vectors (e.g. $\{f_i\}_{i=1}^{N} \subset R^{256}$), strictly corresponding to 3D query boxes.

\textbf{Sparse interaction head.} The sparse interaction head implements local interaction by applying $1 \times 1$ convolutional kernels on RoI features extracted from query boxes and generating corresponding parameters from query features via linear transformation. The inputs are passed through two $1\times 1$ convolutional layers for interaction, followed by a Feed-Forward Network (FFN) block with layer normalization and a linear projection block to output classification and regression predictions. 

\subsection{Sparse Feature Sampling Module} The 3D query boxes, which are learnable, serve as sparse candidates that are updated iteratively. We decode 3D query boxes from center points to box corners through simple geometric transformations. For simplicity, we refer the $i$ th decoded box ${\{C_{il}\}_{l=1}^{8}} \subset \mathbb{R}^3$ with coordinates of eight corners. Through a standard camera model, these query boxes are projected into visible regions of cameras as follows:
\begin{equation}
C_{il}^{*} = C_{il}\oplus \mathrm{1}, C_{mil} = T_m C_{il}^{*},
\end{equation}
where $l =1,\ldots,8, m = 1,\ldots,N_{cam}, C_{mil}=(c_{milx},c_{mily},1)$ and $T_m$ denotes the camera transformation matrix. 
Then the projected boxes on each camera can be obtained as follows:
\begin{equation}
    \tilde{B}_{im}=(\min\limits_{l}c_{imlx},\min\limits_{l}c_{imly}, \max\limits_{l}c_{imlx}, \max\limits_{l}c_{imly}),
    \label{formula_projection}
\end{equation}
where $i = 1,\ldots, N, m = 1,\ldots,N_{cam}$.
\begin{figure}[ht]
\centering
\includegraphics[width=0.45\textwidth]{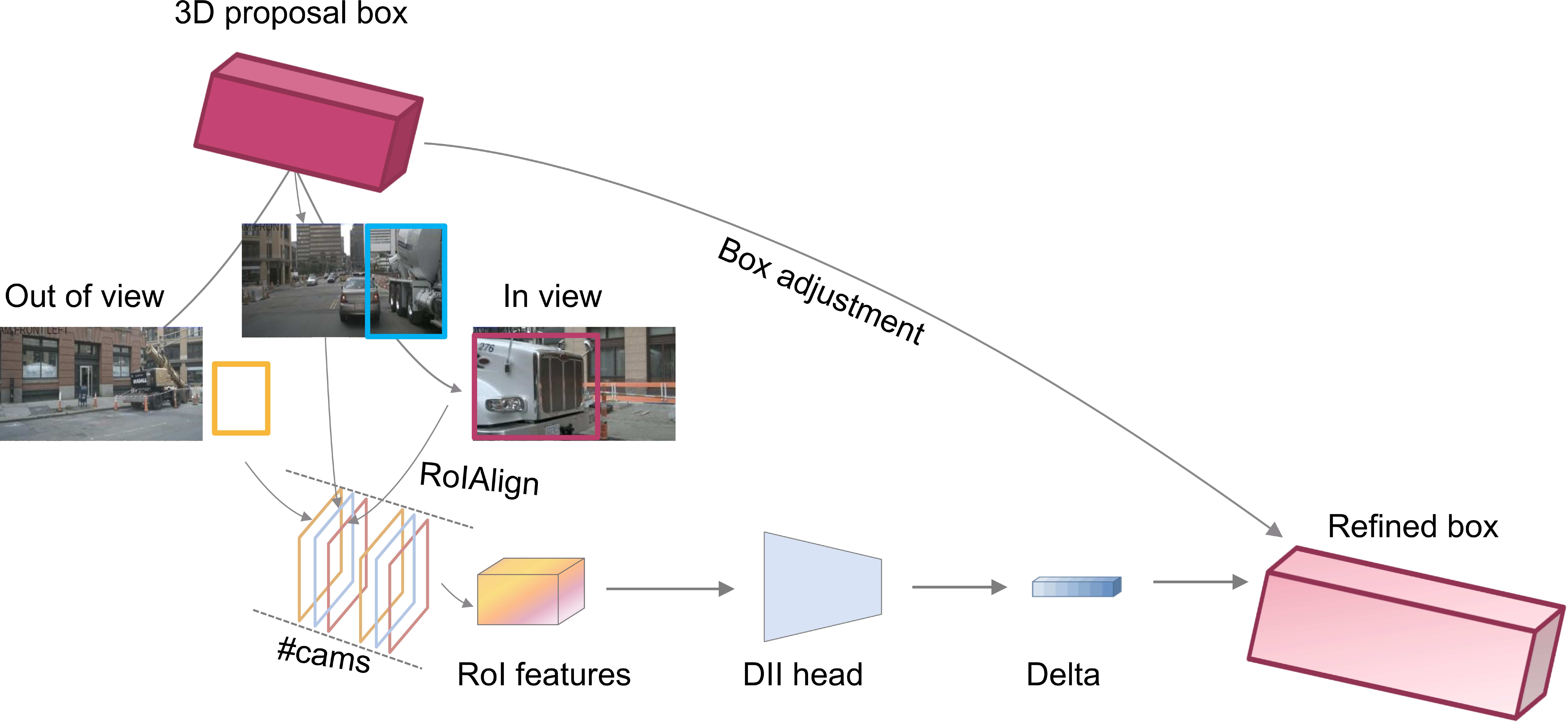}
\caption{A graphical illustration of sparse feature sampling module. 3D query boxes are initialized and projected into feature maps in the camera plane for RoI extraction. After aggregation, RoI features interact with query features in DII head which produces cues for refinement. Finally, box adjustment module refines the boxes.}
\label{sparse_feature_sampling}
\end{figure}

As shown in Fig .\ref{sparse_feature_sampling}, given the projected boxes, we use RoI Align operation to extract features of interest. The projection of box corner points may result in three cases. The normal case indicates a projected 2D box on images. If the projected points have a negative depth, the box locates behind the camera, which is naturally invisible. If the projected corners are outside or partially outside the pixel space, the box is (partially) invisible. The latter two abnormal feature sampling cases result in empty or partially empty RoI features via RoIAlign, so that masking operations and positional encoding techniques are deftly avoided.  

\textbf{Cross-view fusion.} Before entering the prediction head, corresponding RoI features on multi-view camera images are aggregated to guarantee cross-view fusion learning. In this way, RoI features maintain a fixed expression, no matter how many cameras capture one query box. 



\subsection{Box Adjustment}
 We define the predicted boxes and cues in the $t$ th stage as $\{b_i^t\}_{i=1}^{N}$ and $\{\Delta b_i^t\}_{i=1}^{N}$ respectively. Then the box adjustment operation for location and dimension parameters can be formulated as follows:
\begin{equation}
    \left\{\begin{array}{l}
    b_{ix}^{t+1} = \Delta b_{ix}^{t+1} \times b_{iw}^t + b_{ix}^t\\
    b_{iy}^{t+1} = \Delta b_{iy}^{t+1} \times b_{il}^t + b_{iy}^t\\
    b_{iz}^{t+1} = \Delta b_{iz}^{t+1} \times b_{ih}^t + b_{iz}^t\\
    \end{array}\right..
    \label{apply_delta1}
\end{equation}
Considering non-negative constraints, dimension parameters are usually in the form of logarithms. Therefore, its adjustment is formulated as follows:
\begin{equation}
    \left\{\begin{array}{l}
    b_{iw}^{t+1} =  e^{\Delta b_{iw}^{t+1}} \times b_{iw}^t\\
    b_{il}^{t+1} = e^{\Delta b_{il}^{t+1}} \times b_{il}^t\\
    b_{ih}^{t+1} = e^{\Delta b_{ih}^{t+1}} \times b_{ih}^t\\
    \end{array}\right..
    \label{apply_delta2}
\end{equation}
As for rotation and velocity, we directly take the values in $\{\Delta b_i^t\}_{i=1}^{N}$ as the predicted results of the $t$ th stage.

\subsection{Loss Design}
Generally, the loss function of SRCN3D is a linear combination of a Focal Loss \cite{FocalLoss} for category classification and a L1 norm loss for 3D bounding box regression, which is as follows: 
\begin{equation}
    \mathcal{L} = \omega_{cls} \times \mathcal{L}_{cls} +\omega_{l_{reg}} \times \mathcal{L}_{1}.
\end{equation}
SRCN3D employs set prediction loss following DETR \cite{DETR}. Details of set prediction loss are presented in the supplementary material. 


\subsection{Multi-object Tracking}
Multi-object Tracking mainly handles data association of detected objects between past and current frames. Given nonideal detection results, we adopt a hypothesis-oriented probabilistic approach, Multi-Bernoulli Mixture (MBM) to deal with uncertain data association. MBM treats data association into global hypotheses and single target hypotheses. The MBM density is defined as the sum of $j$ global hypotheses as follows:

\begin{equation}
    f^{\mathrm{mbm}}(X)\propto\sum_{j}\sum_{X_1\uplus\ldots\uplus X_n=X}\prod_{i=1}^{n}w_{j,i}f_{j,i}\left(X_i\right),
\end{equation}
where $f^{\mathrm{mbm}}(X)$ is the posterior of MBM intensity, $X$ is the whole set of detected objects, $w_{j,i}$ indicates the weight of a Bernoulli component $X_i$ in global hypothesis $j$, and $f_{j,i}\left(X_i\right)$ is the probability intensity of single Bernoulli component $X_i$ in global hypothesis $j$, defined as follows:
\begin{equation}
  f_{j,i}(X_i)=\left\{\begin{array}{ll}1-r_{j,i} & \textrm{if $X_i=\oslash$}\\
  r_{j,i}p_{j,i}(x) &\textrm{if $X_i={x}$}\\
  0 & \mathrm{\ otherwise\ }
  \end{array}\right.,
\end{equation}
where $r_{j,i}$ denotes the existence probability of object $x$ in single target hypothesis (STH) $X$ and $p_{j,i}$ is the probability density function considering the log likelihood of the hypothesis target. Each STH denotes one object detected or a missed detection. Temporal prediction and update of states follow a standard unscented kalman filter (UKF). An important criterion for matching is to compute likelihood between measurements and states. In SRCN3D tracker, each measurement includes explicit properties of 3D bounding boxes and two kinds of implicit features, namely, RoI features and query features. The likelihood of 3D bounding boxes is computed by Mahanalobis distance introduced in \cite{StanfordTRI}, and latent RoI features and query features follow a cosine similarity. The overall likelihood is calculated as 
\begin{equation}
    l = l_{box} + \alpha \times l_{RoI} + \beta \times l_{prop}.
    \label{motion}
\end{equation}
In practice, we set $\alpha=\beta=0.5$ to achieve a balance among box attributes and appearance.

\section{Experiments}

\begin{table*}[ht] 
    \centering
    \small
        \begin{tabular}{l|cc|cccccccc}
        \toprule
        Method & Size & Backbone &NDS$\uparrow$ & mAP$\uparrow$ & mATE$\downarrow$ & mASE$\downarrow$ & mAOE$\downarrow$ & mAVE$\downarrow$ & mAAE$\downarrow$ & FPS$\uparrow$\\
        \midrule
        CenterNet \cite{centernet} & - & DLA & 0.328 & 0.306 & 0.716 & 0.264 & 0.609 & 1.426 & 0.658 & - \\
        FCOS3D \ddag\#\cite{FCOS3D} & 1600$\times$900 & Res-101 & 0.415 & 0.343 & 0.725 & \textbf{0.263} & 0.422 & 1.292 & \textbf{0.153} & 2.0\\
        DETR3D \P\cite{DETR3D} & 1600$\times$900 & Res-101 & 0.425 & 0.346 & 0.773 & 0.268 & 0.383 & 0.842 & 0.216 & 2.7\\
        BEVDet \S\cite{BEVDET} & 1056$\times$384 & Res-101 & 0.396 & 0.330 & \textbf{0.702} & 0.272 & 0.534 & 0.932 & 0.251 & \textbf{16.7}\\
        BEVFormer-S \P\cite{BEVFormer} & 1600$\times$900 & Res-101 & \textbf{0.448} & \textbf{0.375} & 0.725 & 0.272 & 0.391 & 0.802 & 0.200 & 2.1 \\
        PETR \S\P\cite{PETR} & 1600$\times$900 & Res-101 & 0.442 & 0.370 & 0.711 & 0.267 & 0.383 & 0.865 & 0.201 & 2.5 \\
        SRCN3D (Ours)\P & 1600$\times$900 & Res-101 & 0.428 & 0.337 & 0.779 & 0.287 & \textbf{0.367} & \textbf{0.781} & 0.188 & 3.2\\
        \midrule
        SRCN3D (Ours)\P & 1600$\times$900 & V2-99 & 0.475 & 0.396 & 0.737 & 0.294 & 0.278 & 0.728 & 0.197 & 2.5\\
        \bottomrule
        \end{tabular}
    \caption{Comparison of state-of-the-art detectors on nuScenes detection val set. \ddag: with test-time augmentation. \S: trained with CBGS \protect\cite{CBGS}. \P: initialized from pretrained FCOS3D \protect\cite{FCOS3D} backbone. \#: with model ensemble.}
    \label{val_set_comparison}
\end{table*}

\begin{table*}[htbp]
    \centering
    \setlength{\tabcolsep}{12pt}
    \small
        \begin{tabular}{l|ccccccc}
        \toprule
        Method & NDS$\uparrow$ & mAP$\uparrow$ & mATE$\downarrow$ & mASE$\downarrow$ & mAOE$\downarrow$ & mAVE$\downarrow$ & mAAE$\downarrow$ \\
        \midrule
        CenterNet \cite{centernet} & 0.400 & 0.338 & 0.658 & 0.255 & 0.629 & 1.629 & 0.142 \\
        EPro-PnP-Det\cite{Epro-PnP-Det} & 0.453 & 0.373 & 0.605 & 0.243 & \textbf{0.359} & 1.067 & \textbf{0.124} \\
        M2BEV\cite{M2BEV} & 0.451 & 0.398 & 0.577 & 0.245 & 0.500 & 1.227 & 0.154\\
        FCOS3D \ddag\cite{FCOS3D} & 0.428 & 0.358 & 0.690 & 0.249 & 0.452 & 1.434 & \textbf{0.124} \\
        BEVFormer-S \cite{BEVFormer} &  0.462 & 0.409 & 0.650 & 0.261 & 0.439 & 0.925 & 0.147 \\
        DETR3D \dag\cite{DETR3D} &  0.479 & 0.412 & 0.641 & 0.255 & 0.394 & 0.845 & 0.133\\
        DD3D \dag\ddag\cite{DD3D} &  0.477 & 0.418 & 0.572 & 0.249 & 0.368 & 1.014 & 0.124 \\
        BEVDet \dag\cite{BEVDET} & 0.488 & 0.424 & \textbf{0.524} & \textbf{0.242} & 0.373 & 0.950 & 0.148 \\
        PETR \dag\cite{PETR} &  \textbf{0.504} & \textbf{0.441} & 0.593 & 0.249 & 0.383 & \textbf{0.808} & 0.132 \\
        SRCN3D \dag(Ours) &  0.463 & 0.396 & 0.673 & 0.269 & 0.403 & 0.875 & 0.129\\
        \bottomrule
        \end{tabular}
    \caption{Comparison of state-of-the-art detectors on nuScenes detection test set. Detection methods using temporal aggregation are not included. \dag: trained using extra data. \ddag: with test time augmentation.}
    \label{test_set_comparison}
\end{table*}
\subsection{Dataset} We report experiment results on large-scale public nuScenes dataset \cite{nuscenes}, which includes 1000 driving scenes of about 20 seconds duration. RGB images are collected from 6 cameras with known intrinsic and extrinsic camera parameters. NuScenes dataset provides $28130$, $6019$ and $6008$ samples for training, validation and testing, respectively. Only key frames at $2$Hz are annotated and used. 


\subsection{Metrics} We adopt the nuScenes \cite{nuscenes} official evaluation protocol. As for detection metrics, we adopt mean average precision (mAP) and nuScenes Detection Score (NDS) as primary metrics, and true positive metrics (TP metrics) including average translation error (ATE), average scale error (ASE), average orientation error (AOE), average velocity error (AVE), and average attribute error (AAE). Metrics of MOT are based on CLEARMOT \cite{CLEARMOT}, including average multi object tracking accuracy (AMOTA) as the primary metric, and average multi-object tracking precision (AMOTP) and recall rate (RECALL). Reports of SRCN3D on all detection and tracking metrics are publicly available on nuScenes leaderboard.

\subsection{Training and Inference}

\textbf{Training.} Our code is built upon the MMDetection3D \cite{mmdet3d2020}. We employ two types of backbone network: (1) ResNet101 \cite{Resnet} pretrained on FCOS3D \cite{FCOS3D} and PGD \cite{PGD}, (2) VoVNetV2-99 \cite{V2-99} pretrained on DD3D \cite{DD3D}. We use a bipartite loss via 3D Hungarian assigner, which consists of a Focal Loss \cite{FocalLoss} with weight $2.0$ and a L1 loss with weight $0.25$. The model is trained using AdamW\cite{AdamW} optimizer with weight decay of $0.01$. The learning rate is initialized with $2e^{-4}$ and decays with cosine annealing policy. We set the detection region to $[-61.2m, 61.2m]$ for the $X$ and $Y$ axis, and $[-5m, 3m]$ for the $Z$ axis. Experiments are trained for 24 epochs on 8 2080TI GPUs, the training hours is around 24 hours.


\textbf{Inference.} In the inference process, SRCN3D makes simple predictions without any post-processing step such as NMS and test-time augmentation (TTA). Our model infers within $3.2$ FPS on a single RTX3090 GPU with ResNet 101 as the backbone and the original image size $1600\times900$ as shown in Table. \ref{val_set_comparison}, faster than other transformer-based detectors like DETR3D \cite{DETR3D} with sparse queries ($2.7$ FPS), BEVFormer \cite{BEVFormer} ($2.1$ FPS) with dense queries and PETR \cite{PETR} with hybrid queries.

\subsection{Comparison with State-of-the-art}

\begin{table*}[htbp]
    \centering
    \small
    \setlength{\tabcolsep}{11pt}
    \begin{tabular}{l|ll|ccc}
        \toprule
        Method & Modality & Split & AMOTA$\uparrow$ & AMOTP$\downarrow$ & RECALL$\uparrow$ \\
        \midrule
        QD3DT\cite{QD-3DT} & C & val & 0.242 & 1.518 & 0.399\\
        MUTR3D\cite{MUTR3D} & C & val & 0.294 & 1.498 & 0.427\\
        ViP3D\cite{ViP3D} & C & val & 0.216 & 1.616 & 0.358\\
        UniAD\cite{UniAD} & C & val & 0.359 & 1.320 & 0.467 \\ 
        SRCN3D (Ours) & C & val & \textbf{0.439} & \textbf{1.280} & \textbf{0.545}\\
        \midrule
        CenterTrack-Open \cite{CenterTrack} & L + C & test & 0.108 & \textbf{0.989} & 0.412 \\
        QD-3DT \cite{QD-3DT} & C & test & 0.217 & 1.550 & 0.375\\
        PolarDETR \cite{PolarDETR} & C & test & 0.273 & 1.185 & 0.404\\
        DEFT \cite{DEFT} & C & test & 0.177 & 1.564 & 0.338 \\
        MUTR3D\cite{MUTR3D} & C & test & 0.270 & 1.494 & 0.411\\
        SRCN3D (Ours) & C & test & \textbf{0.398} & 1.317 & \textbf{0.538}\\ 
        \bottomrule
    \end{tabular}
    \caption{Comparison of state-of-the-art trackers on nuScenes tracking benchmark. The quantitative results for the validation set are obtained from the original paper, while the results for the test set are obtained from the nuScenes leaderboard. For modalities, “L” denotes LiDAR and “C” denotes camera.}
    \label{test_comparison_tracking}
\end{table*}

\textbf{nuScenes detection benchmark.} In Table. \ref{val_set_comparison}, we present the performance comparison with state-of-the-art methods on nuScenes validation set. SRCN3D gains 42.8\% NDS and 33.8\% mAP for ResNet101 \cite{Resnet} backbone and image size of $900\times 1600$. Compared to Monocular 3D detectors, SRCN3D surpasses CenterNet \cite{centernet} and FCOS3D \cite{FCOS3D} in NDS by 10.1\% and 1.3\%. Compared to MV3D detectors, SRCN3D also shows competitive results. Under the same backbone and image size settings, our method 
outperforms DETR3D \cite{DETR3D} by 0.3\% in NDS. In terms of TP metrics, it shows that SRCN3D works well in predicting orientation, velocity and attributes, achieving the best performance in mAOE and mAVE. However, SRCN3D still suffer from limitations for translation and scale predictions. Table. \ref{test_set_comparison} shows the performance comparison on nuScenes detection test set. Our method achieves competitive performance on NDS, mAP and other true positive metrics. Overall, the experimental results demonstrate the effectiveness of our method on 3D object detection tasks.

\begin{table}[!ht]
    \centering
    \small
    \setlength{\tabcolsep}{6pt}
    \begin{tabular}{ccccc|cc}
    \toprule
    Box & Head & Delta & Init  & Others & NDS$\uparrow$   & mAP$\uparrow$   \\
    \midrule
    \checkmark &       &             &      Random    &  - &    0.310 & 0.249 \\
              &   \checkmark    &             &          Random  &  - &   0.341 & 0.258 \\
    \checkmark &   \checkmark    &             &    Random  &  - &       0.409 & 0.333 \\
    \checkmark &   \checkmark    &      \checkmark       &   Fixed  &  - &         0.414 & 0.329 \\
    \checkmark &   \checkmark    &      \checkmark       &   Random  &  - &        \textbf{0.428} & 0.337 \\
    \checkmark &   \checkmark    &      \checkmark       &   Random  &  Self-attn &   0.427 & \textbf{0.338} \\
    \bottomrule
    \end{tabular}
    \caption{Ablation on key modules. “Box” denotes the two-stage query boxes module. “Head” denotes the sparse interaction head. “Delta” denotes the box adjustment module. “Init” denotes the initialization method for query boxes. “Self-attn” means we add a self-attention module to the feature sampling module.}
    \label{ablation_modules}
\end{table}

\textbf{nuScenes tracking benchmark.} Table. \ref{test_comparison_tracking} reports nuScenes tracking benchmark on both validation and test split. SRCN3D achieves state-of-the-art performance in camera track and exceeds other competitors by a large margin. On validation set, compared with camera based methods, SRCN3D achieves the best performance in all reported metrics. On test set, our method achieves $0.398$ in terms of AMOTA metrics on nuScenes test set, more than $12$ points of accuracy improvement over recent state-of-the-art camera-only trackers. The test results also show a moderate AMOTP error and the highest recall rate.

\subsection{Ablation Study}

\begin{table}[h!]
\centering
 \small
\setlength{\tabcolsep}{10pt}
\begin{tabular}{l|ccc}
\toprule
Modules & AMOTA$\uparrow$ & AMOTP$\downarrow$ & RECALL$\uparrow$ \\
\midrule
DE    &    0.277  & 1.519      & 0.506             \\
PR     &       0.405&       1.361&    0.508  \\
PR + R    &       0.436&       1.287&    0.539  \\
PR + H   &       \textbf{0.439}&       \textbf{1.280}&     \textbf{0.545}   \\
\bottomrule
\end{tabular}
\caption{Ablation of the tracking module. Deterministic method is our adaption of AB3DMOT \cite{AB3DMOT}, The other lines shows utilization of different features in data association process. “DE” denotes deterministic matching method. “PR” denotes probabilistic matching method. “R”, and “H” denotes RoI features and hybrid features, respectively.}
\label{tab_ablation_tracking}
\end{table}

\begin{figure*}[ht]
	\centering
	\includegraphics[width=1 \textwidth]{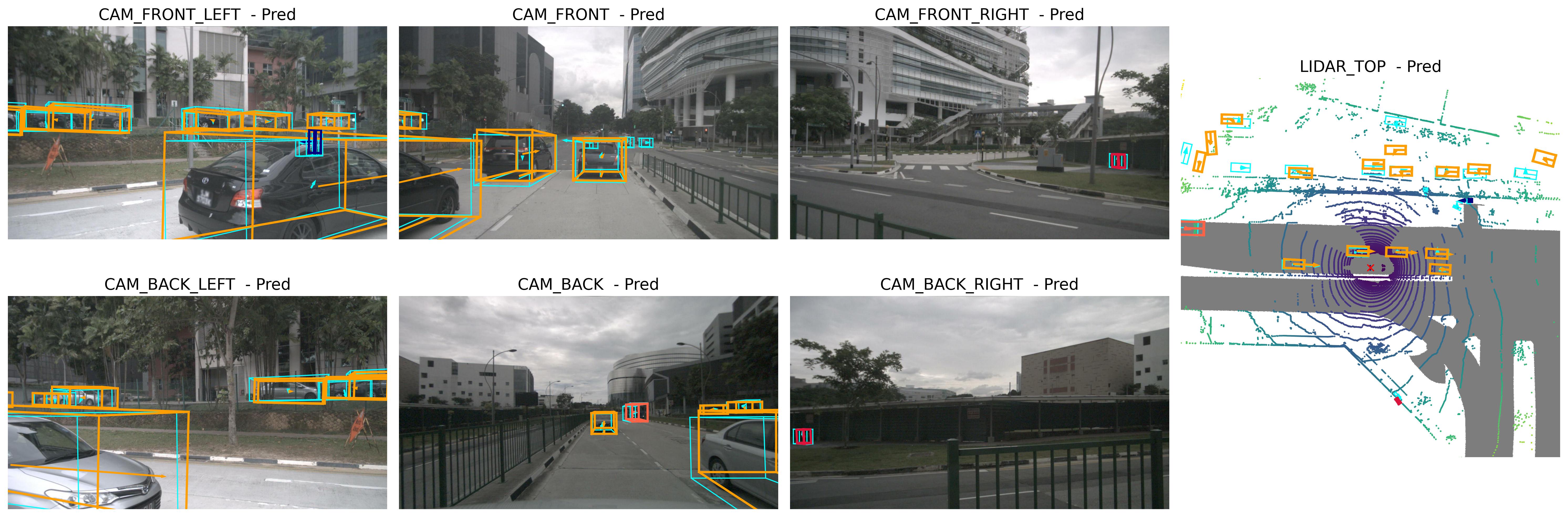}
	\caption{Visualization of final predictions and ground truth boxes on nuScenes val set. Ground truth boxes are coloured in light green. Boxes coloured in orange, red and blue correspond to vehicles, bicycles and pedestrians, respectively.}
	\label{prediction}
\end{figure*}

In this section, we perform ablations on several important components or properties of SRCN3D on nuScenes validation set. 


\begin{table}[h!]
    \centering
  \small
    \begin{tabular}{cccc|cc}
    \toprule
    Backbone & Queries & Objects & Stages & NDS$\uparrow$   & mAP$\uparrow$    \\
    \midrule
    R-101 & 900   & 300 & 6 & 0.428 & 0.337 \\
    R-101 & 500   & 300 & 6 & 0.408 & 0.332 \\
    R-101 & 300   & 300 & 6 & 0.418 & 0.327 \\
    V2-99 & 900   & 300 & 6 & \textbf{0.475} & \textbf{0.396} \\
    V2-99 & 500   & 300 & 6 & 0.474 & 0.378 \\     
    \midrule
    R-101 & 300   & 300 & 6 & \textbf{0.418} & \textbf{0.327} \\
    R-101 & 300   & 200 & 6 & 0.410 & 0.324 \\
    \midrule
    R-101 & 900   & 300 & 6 & \textbf{0.428} & \textbf{0.337} \\
    R-101 & 900   & 300 & 3 & 0.421 & 0.335 \\
    R-101 & 900   & 300 & 1 & 0.385 & 0.297 \\   

    \bottomrule
    \end{tabular}
    \caption{Ablation on the number of queries, candidate objects and stages.}
    \label{ablation_num_object_stage}
\end{table}

\textbf{Ablation on key modules.} We have identified four key modules that distinguish SRCN3D from previous approaches: learnable boxes, sparse interaction module, box adjustment, and initialization. To evaluate the importance of these modules, we perform ablation experiments by replacing them with similar parts in DETR3D. Table \ref{ablation_modules} shows the results. In the first case, we construct a two-stage DETR3D, showing that query features alone are not sufficient for refining query boxes. The second case regresses boxes directly using a regression branch of query features at the beginning. The third case removes the box adjustment step and directly inputs object features into the regression branch, resulting in a slightly reduced accuracy. Additionally, random initialization of query boxes outperforms fixed initialization. We also test the dynamic instance interaction head introduced in Sparse R-CNN, which has an additional self-attention module. Our sparse interaction without self-attention is equally effective, demonstrating that we do not need self-attention to distinguish different queries.

\textbf{Ablation on number of queries, objects and stages.} The number of query boxes and features is related to GPU memory cost and inference speed. Number of objects determines the output of NMS-free box coder. Number of stages refers to how many times the query boxes are refined. 

\begin{figure}
	\centering
	\includegraphics[width=0.25 \textwidth]{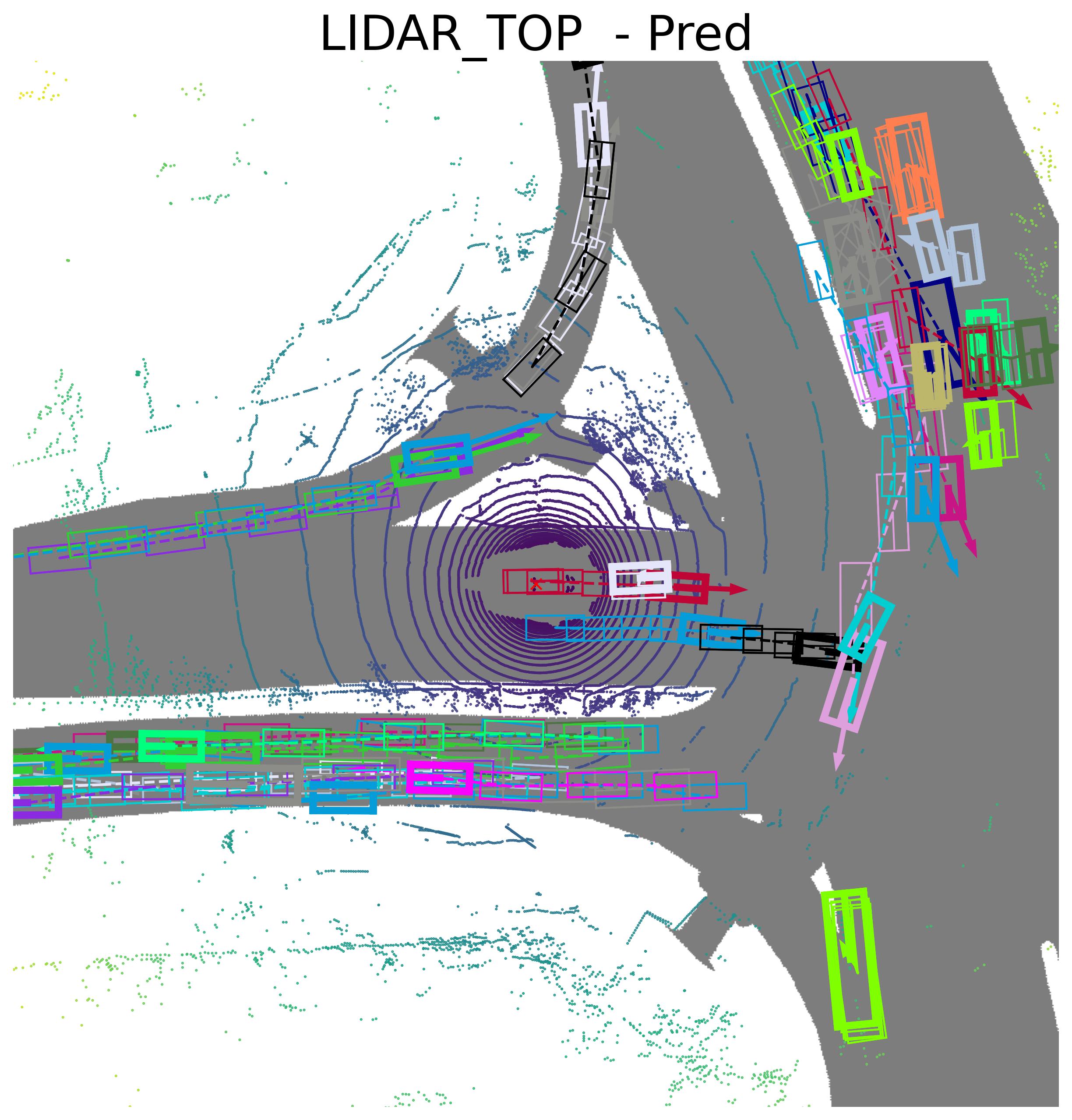}
	\caption{Visualization of tracking results on nuScenes val set. We visualize the tracked objects in the past five key frames in the same scene. Boxes in different colors refer to different tracked instances.}
	\label{tracking}
\end{figure}

Table \ref{ablation_num_object_stage} shows that reducing the number of queries has only a slight impact on accuracy ($<3\%$), while reducing the number of candidate objects leads to a greater loss of accuracy. Additionally, the first three stages in the cascade structure contribute significantly to improving accuracy, while later stages provide diminishing returns. This suggests that the refinement process efficiently converges by the third stage. Our ablation study demonstrates that our sparse feature sampling module can capture regions of interest using only local features in a lightweight manner.

\textbf{Ablation on tracking module.} As shown in Table. \ref{tab_ablation_tracking}, for camera-only tracking, probabilistic association greatly outperforms deterministic counterparts. We further demonstrate that as RoI features and query features represent kinds of affinity cues for appearance and classification respectively, these latent features facilitate the matching process and improve tracking precision. 

\subsection{Visualization}
Fig. \ref{prediction} visualizes final detection results in the camera front view and bird's-eye view on the validation set with ground truth annotation. Overall, as shown in the bird's-eye view, the predicted boxes are close to the ground truth ones. The bus detected as a whole both in front and left-front camera illustrates the effectiveness of cross-view fusion in overlapped regions. Small objects (e.g. pedestrians in the front-left camera view) are also detected precisely. These results indicate a satisfactory performance of SRCN3D and related modules. More visualizations of final results and 
intermediate-stage query boxes are available in the supplementary materials.

Fig. \ref{tracking} presents a BEV example of tracking on nuScenes validation set. We visualize past five key frames of unique objects in a crowded intersection to demonstrate satisfactory tracking accuracy and continuity.

Visualization also exposes current limitations of SRCN3D. In areas where features of targeted objects are dense, the predicted boxes overlap with each other, which is unreasonable for real-world objects. It shows that there remains a few duplicates in predicted boxes.

\subsection{Discussion}
Sparse R-CNN 3D provides the first attempt for box-wise sampling and refinement approach to conduct 3D object detection and downstream multi-object tracking. The RoI features are commonly used in two-stage object detection pipelines, serving as a downstream refinement of region proposal network (RPN). In contrast to grid-sampled pixel-level feature sampling, RoI features offer a comprehensive perspective of the region, making them better suited for tasks such as classification, orientation, and velocity estimation. In contrast, RoI features are not as effective as grid sampling in object localization tasks in 3D space, which leads to larger errors in object center and scale estimation.

\section{Conclusion}
This paper proposes a novel innovative architecture, SRCN3D, aiming at detecting and tracking objects of interest. It possesses the traits of sparse queries, sparse attention and sparse prediction, and is able to efficiently extract and fuse cross-view features. Our insight is that box-feature twin-track queries and cascade-style refinement process with only local RoI attention enable 3D object detection and cross-view fusion. We hope that this architecture can serve as a foundation for fully-sparse surround-view 3D object detection. In the future, the authors will investigate deeper in combining segmentation and temporal information to enhance the accuracy and robustness of SRCN3D.

\section*{Acknowledgement}
This work was supported in part by the National Natural Science Foundation of China under Grants U22A20104, and Beijing Municipal Science and Technology Commission (Grant No.Z221100008122011).

{\small
\bibliographystyle{ieee_fullname}
\bibliography{egbib}
}

\end{document}